\documentclass[twocolumn]{IEEEtran}
\usepackage[dvips]{graphicx}
\usepackage{subeqn}
\newcommand{\gbf}[1] {\mbox{\boldmath${#1}$\unboldmath}}
\newcommand{\be}{\begin{equation}}
\newcommand{\ee}{\end{equation}}
\newcommand{\beq}{\begin{equation}}
\newcommand{\eeq}{\end{equation}}
\newcommand{\bed}{\begin{displaymath}}
\newcommand{\eed}{\end{displaymath}}
\newcommand{\beqa}{\begin{eqnarray}}
\newcommand{\eeqa}{\end{eqnarray}}
\newcommand{\beqann}{\begin{eqnarray*}}
\newcommand{\eeqann}{\end{eqnarray*}}
\newcommand{\bseq}{\begin{subequations}}
\newcommand{\eseq}{\end{subequations}}

\newcommand{\ba}{\begin{array}}
\newcommand{\ea}{\end{array}}

\newcommand{\negr}[1]{{\bf {#1}}}

\begin{document}
\title {Architecture Optimization of \goodbreak
a 3-DOF Translational Parallel Mechanism for \goodbreak Machining
Applications, the Orthoglide}
\author{Damien Chablat, Philippe Wenger (corresponding author) \thanks{Damien Chablat and Philippe Wenger are
with the Institut de Recherche en Communications et Cybern\'etique
de Nantes (IRCCyN), 1, rue de la No\"e, 44321 Nantes, France,
email: Philippe.Wenger@irccyn.ec-nantes.fr}} \markboth{W2001-255
Regular paper accepted to IEEE Transactions On Robotics and
Automation} {Damien Chablat, Philippe Wenger} \maketitle
\begin{abstract}
This paper addresses the architecture optimization of a 3-DOF
translational parallel mechanism designed for machining
applications. The design optimization is conducted on the basis of
a prescribed Cartesian workspace with prescribed kinetostatic
performances. The resulting machine, the Orthoglide, features
three fixed parallel linear joints which are mounted orthogonally
and a mobile platform which moves in the Cartesian $x$-$y$-$z$
space with fixed orientation. The interesting features of the
Orthoglide are a regular Cartesian workspace shape, uniform
performances in all directions and good compactness. A small-scale
prototype of the Orthoglide under development is presented at the
end of this paper.
\end{abstract}
\section{Introduction}
\PARstart{P}{arallel} kinematic machines (PKM) are commonly
claimed to offer several advantages over their serial
counterparts, like high structural rigidity, high dynamic
capacities and high accuracy \cite{Tlusty:1999,Wenger:1999a}.
Thus, PKM are interesting alternative designs for high-speed
machining applications.

This is why parallel kinematic machine-tools attract the interest
of more and more researchers and companies. Since the first
prototype presented in 1994 during the IMTS in Chicago by
Gidding\&Lewis (the VARIAX), many other prototypes have appeared.

However, the existing PKM suffer from two major drawbacks, namely,
a complex workspace and highly non linear input/output relations.
For most PKM, the Jacobian matrix which relates the joint rates to
the output velocities is not constant and not isotropic.
Consequently, the performances {\it e.g.} maximum speeds, forces,
accuracy and rigidity) vary considerably for different points in
the Cartesian workspace and for different directions at one given
point. This is a serious drawback for machining applications
\cite{Tlusty:1999,Kim:1997,Wenger:1999b}. To be of interest for
machining applications, a PKM should preserve good workspace
properties, that is, regular shape and acceptable kinetostatic
performances throughout. In milling applications, the machining
conditions must remain constant along the whole tool path
\cite{Rehsteiner:1999a}. In many research papers, this criterion
is not taken into account in the algorithmic methods used for the
optimization of the workspace volume \cite{Luh:1996,Merlet:1999}.

Most industrial 3-axis machine-tools have a serial kinematic
architecture with orthogonal linear joint axes along the x, y and
z directions. Thus, the motion of the tool in any of these
directions is linearly related to the motion of one of the three
actuated axes. Also, the performances are constant throughout the
Cartesian workspace, which is a parallelepiped. The main drawback
is inherent to the serial arrangement of the links, namely, poor
dynamic performances. The purpose of this paper is to design a
translational $3$-axis PKM with the advantages of serial machine
tools but without their drawbacks. Starting from a Delta-type
architecture with three fixed linear joints and three articulated
parallelograms, an optimization procedure is conducted in which
two criteria are used successively, (i) the conditioning of the
Jacobian matrix of the PKM
\cite{Golub:1989,Salisbury:1982,Huang:1998,Zanganeh:1998} and (ii)
the manipulability ellipsoid \cite{Yoshikawa:1985}. The first
criterion leads to an isotropic architecture that features a
configuration where the tool forces and velocities are equal in
all directions. The second criterion makes it possible to define
the actuated joint limits and the link lengths with respect to a
desired Cartesian workspace size and prescribed limits on the
transmission factors. The resulting PKM, the Orthoglide, has a
Cartesian workspace shape that is close to a cube whose sides are
parallel to the planes $xy$, $yz$ and $xz$ respectively. A
systematic design procedure is proposed to define the geometric
parameters as function of the size of a prescribed cubic Cartesian
workspace and bounded velocity and force transmission factors
throughout.

Next section presents the existing PKM. The design parameters and
the kinematics of the mechanism to be optimized are reported in
Section~3. Section~4 is devoted to the design procedure of the
Orthoglide and the presentation of the prototype.
\section{Existing PKM}
Most existing PKM can be classified into two main families. The
PKM of the first family have fixed foot points and variable length
struts. These PKM are generally called ``hexapods'' when they have
6 degrees of freedom. Hexapods have a Stewart-Gough parallel
kinematic architecture. Many prototypes and commercial hexapod PKM
already exist like the VARIAX (Gidding\&Lewis), the CMW300
(Compagnie M\'ecanique des Vosges), the TORNADO 2000 (Hexel), the
MIKROMAT 6X (Mikromat/IWU), the hexapod OKUMA (Okuma), the hexapod
G500 (GEODETIC). In this first family, we find also hybrid
architectures with a 2-axis wrist mounted in series to a 3-DOF
``tripod'' positioning structure ({\it e.g.} the TRICEPT from
Neos-Robotics \cite{Tricept:1988}). Since many machining tasks
require only 3 translational degrees of freedom, several 3-axis
translational PKM have been proposed. There are several ways to
design such mechanisms
\cite{Reboulet:1990,Herve:1991,Carricato:2002,Kong:2002}. In the
first family, we find the Tsai mechanism and its variants. In
these mechanisms, the mobile platform is connected to the base by
three extensible limbs with a special arrangement of the universal
joints that restrains completely the orientation of the mobile
platform \cite{Tsai:2000,Digregorio:1998}.

The PKM of the second family have fixed length struts with
moveable foot points gliding on fixed linear joints. In this
category we find the HEXAGLIDE (ETH Z\"{u}rich) which features six
parallel (also in the geometrical sense) and coplanar linear
joints. The HexaM (Toyoda) is another example with three pairs of
adjacent linear joints lying on a vertical cone
\cite{patent:1998}. A hybrid parallel/kinematic PKM with three
inclined linear joints and a two-axis wrist is the GEORGE V (IFW
Uni Hanover). Many 3-axis translational PKM belong to this second
family and use an architecture close to the linear Delta robot
originally designed by Reymond Clavel for pick-and-place
operations \cite{Clavel:1988}. In this architecture, three
parallelograms are used to provide the moving platform with pure
translations. The TRIGLIDE (Mikron) has three parallel linear
joints in an horizontal plane. The LINAPOD and the INDEX V100 have
three vertical (non coplanar) linear joints \cite{Pritschow:1997}.
The Urane SX (Renault Automation) and the QUICKSTEP (Krause \&
Mauser) have with three non coplanar horizontal linear joints
\cite{Company:2000}. The aforementioned five machines have
parallel linear joints. This feature provides these machines with
high stiffness in the direction of the linear joints and poor
stiffness in the orthogonal directions. Thus, these machines are
more suitable for specialized operations like drilling, than for
general machining tasks. The STAR mechanism has three horizontal
linear joints intersecting at one point \cite{Herve:1991}.
Isotropic conditions for the STAR mechanisms were studied in
\cite{Baron:2001} but a special type of singularity was shown to
occur at the isotropic configuration if one prescribes unitary
transmission factors \cite{Majou:2002}. At this singularity (a
so-called ``RPM-IO-II singularity''  in the classification of
\cite{Zlatanov:1994}), there is a loss of both input and output
motions and, at the same time, a redundant passive motion of each
leg occurs. Recently, one 3-DOF translational mechanism with
gliding foot points was found in three separate works to be
isotropic throughout the Cartesian workspace
\cite{Carricato:2002,Kong:2002,Kim:2002}. The mobile platform is
connected to three orthogonal linear drives through three
identical planar 3-revolute jointed serial chains. Full isotropy
is clearly an outstanding property. On the other hand, bulky legs
are required to assure stiffness because these legs are subject to
bending.

PKM with fixed length struts and moveable foot points are
interesting because the actuators are fixed and the moving masses
are lower than in the hexapods and tripods.

\section{Problem formulation}
\subsection{Design Parameters}
The machine-tool we want to design is a spatial translational PKM
dedicated to general $3$-axis machining tasks with the following
requirements, (i) a configuration should exist where the
transmission factors are equal to one in all directions, like in a
translational serial machine (ii) the Cartesian workspace shape
should be close to a cube of prescribed size with regular
performances throughout, (iii) the design should be symmetric and
use simple joints to lower the manufacturing costs, (iv) the PKM
should be intrinsically stiff and (v) the PKM should have fixed
linear actuated joints to lower the moving masses. To meet the
last requirement, we start with a PKM architecture of the second
family {\it i.e.} with fixed linear joints. The use of three
articulated parallelograms assembled in an over-constrained way is
an interesting solution to comply with requirement (iv).
Requirements (i) and (ii) will be satisfied in Section~4 by the
isotropic conditions and limited transmission factors constraints.
It will be shown that requirement (i) imposes that the three
actuated linear joint must be orthogonal, hence the name
``orthoglide''. To fulfill requirement (iii), finally, the three
legs should use only revolute joints and be identical.

\begin{figure}[!ht]
  \begin{center}
    \includegraphics[width=60mm,height=42mm]{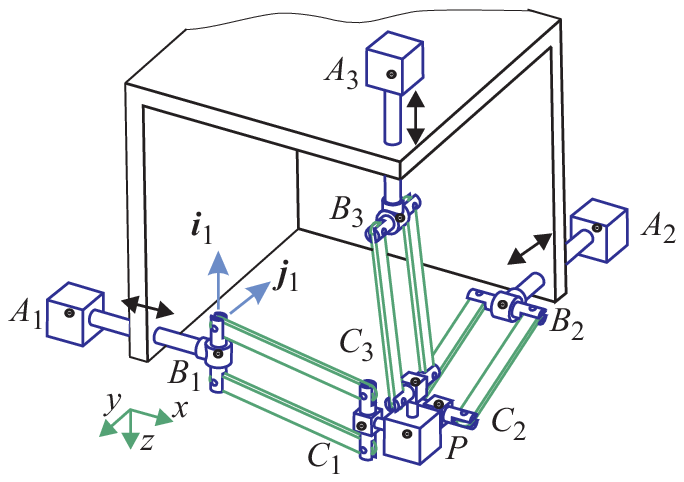}
    \caption{Basic kinematic architecture}
    \protect\label{figure:Orthoglide}
  \end{center}
\end{figure}

Figure~\ref{figure:Orthoglide} shows the basic kinematic
architecture of a PKM that complies with requirements (iii), (iv)
and (v) and that we will optimize with respect to requirements (i)
and (ii). For more simplicity, the figure shows the PKM with the
optimized ({\it i.e.} orthogonal) linear joints arrangement.

\begin{figure}[!ht]
    \begin{center}
           \includegraphics[width=40mm,height=32mm]{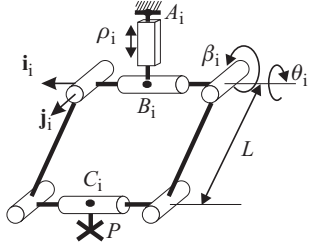}
           \caption{Leg kinematics}
           \protect\label{figure:Leg_Kinematic}
    \end{center}
\end{figure}

The linear joints can be actuated by means of linear motors or by
conventional rotary motors with ball screws. Like in the
Delta-type PKM, the output body is connected to the linear joints
through a set of three parallelograms of equal lengths
$L~=~B_iC_i$, so that it can move only in translation. The three
legs are $PRPaR$ identical chains, where $P$, $R$ and $Pa$ stands
for Prismatic, Revolute and Parallelogram joint, respectively.
Thus, the mechanism is over-constrained. The arrangement of the
joints in the $PRPaR$ chains have been defined to eliminate any
special singularity \cite{Majou:2002}. Each base point $A_i$ is
fixed on the $i^{th}$ linear axis such that
$A_1A_2=~A_1A_3=~A_2A_3$. The points $B_i$ and $C_i$ are located
on the $i^{th}$ parallelogram as shown in
Fig.~\ref{figure:Leg_Kinematic}.

The design parameters to be optimized are the parallelogram
length, the position and orientation of each linear actuated joint
axis and the range of the linear actuators.

\subsection{Kinematic Equations and Singular configurations}
Let $\theta_i$ and $\beta_i$ denote the joint angles of the
parallelogram about the axes $\negr i_i$ and $\negr j_i$,
respectively (Fig.~\ref{figure:Leg_Kinematic}). Let $\negr
\rho_1$, $\negr \rho_2$, $\negr \rho_3$ denote the linear joint
variables, $\negr \rho_i=A_iB_i$.

Let $\dot{\gbf {\rho}}$ be referred to as the vector of actuated
joint rates and $\dot{\negr p}$ as the velocity vector of point
$P$:
 \bed
    \dot{\gbf{\rho}}=
    [\dot{\rho}_1~\dot{\rho}_2~\dot{\rho}_3]^T
   ,\quad
    \dot{\negr p}=
    [\dot{x}~\dot{y}~\dot{z}]^T
 \eed
$\dot{\negr p}$ can be written in three different ways by
traversing the three chains $A_iB_iC_iP$:
 \be
    \dot{\negr p} =
    \negr n_i \dot{\rho}_i +
    (\dot{\theta}_i \negr i_i + \dot{\beta}_i \negr j_i)
    \times
    (\negr c_i - \negr b_i)
 \label{equation:cinematique}
 \ee
where $\negr b_i$ and $\negr c_i$ are the position vectors, in a
given reference frame, of the points $B_i$ and $C_i$,
respectively, and $\negr n_i$ is the direction vector of the
linear joints, for $i=1, 2, $3.

We want to eliminate the two passive joint rates $\dot{\theta}_i$
and $\dot{\beta}_i$ from Eqs.~(\ref{equation:cinematique}), which
we do upon dot-multiplying Eqs.~(\ref{equation:cinematique}) by
$\negr c_i - \negr b_i$:
 \be
   (\negr c_i - \negr b_i)^T \dot{\negr p} =
   (\negr c_i - \negr b_i)^T
   \negr n_i \dot{\rho}_i
 \label{equation:cinematique-2}
 \ee
Equations (\ref{equation:cinematique-2}) can now be cast in vector
form, namely
 \bed
   \negr A \dot{\bf p} = \negr B \dot{\gbf \rho}
 \eed
where \negr A and \negr B are the parallel and serial Jacobian
matrices, respectively:
 \bseq
 \beqa
   \negr A =
   \left[\begin{array}{c}
           (\negr c_1 - \negr b_1)^T \\
           (\negr c_2 - \negr b_2)^T \\
           (\negr c_3 - \negr b_3)^T
         \end{array}
   \right] \\
   \negr B =
   \left[\begin{array}{ccc}
            \eta_1&
            0 &
            0 \\
            0 &
            \eta_2&
            0 \\
            0 &
            0 &
            \eta_3
         \end{array}
   \right]
 \eeqa
 \label{equation:A_et_B}
 \eseq
with $\eta_i= (\negr c_i - \negr b_i)^T \negr n_i $ for $i =
1,2,3$.

\begin{figure}[!ht]
    \begin{center}
           \includegraphics[width=65mm,height=53mm]{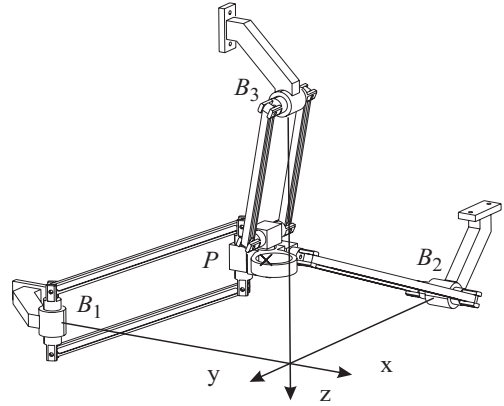}
           \caption{Parallel singular configuration in the general case}
           \protect\label{figure:Coplanarsing}
    \end{center}
\end{figure}

\begin{figure}[!ht]
    \begin{center}
           \includegraphics[width=34mm,height=42mm]{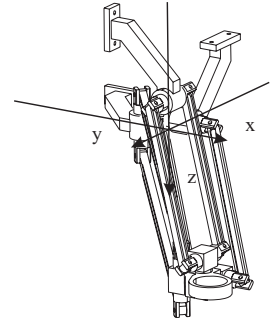}
           \caption{Parallel singular configuration when $B_iC_i$ are parallel}
           \protect\label{figure:Parallelsing}
    \end{center}
\end{figure}

The parallel singularities occur when the determinant of the
matrix \negr A vanishes, {\it i.e.} when $\det(\negr A)=0$. In
such configurations, it is possible to move locally the mobile
platform whereas the actuated joints are locked. These
singularities are particularly undesirable because the structure
cannot resist any force. Equation~(\ref{equation:A_et_B}a) shows
that the parallel singularities occur when the three vectors
$\negr c_i - \negr b_i$ are linearly dependent, that is when the
pairs of points ($B_i$, $C_i$) lie in parallel planes
(Fig.~\ref{figure:Coplanarsing}). To interpret this singularity,
it is more convenient to regard the points $C_i$ as coincident
(this does not change the analysis since each offset $C_iP$ can be
included in $\rho_i$). Then, a parallel singularity occurs when
the points $B_1$, $B_2$, $B_3$ and $C=C_1=C_2=C_3=P$ are coplanar.
Since, at a parallel singular configuration, $P$ is always equally
distant from $B_1$, $B_2$ and $B_3$, $P$ is at the center of a
circle of radius $L$ that cuts the $x$, $y$ and $z$ axes at $B_1$,
$B_2$ and $B_3$, respectively, where $x$, $y$ and $z$ are parallel
to the three linear actuated joints, respectively
(Fig.~\ref{figure:Coplanarsing}). The parallel singularities are
defined by the surface generated by $P$ when this circle
``glides'' along the $x$, $y$ and $z$ axes. A particular parallel
singularity occurs when the links $B_iC_i$ are parallel. The
surface generated is a sphere of radius $L$ and centered at the
intersection of the $x$, $y$ and $z$ axes
(Fig.~\ref{figure:Parallelsing}).

Serial singularities arise when the serial Jacobian matrix \negr B
is no longer invertible {\it i.e.} when $\det(\negr B)=0$. At a
serial singularity a direction exists along which any Cartesian
velocity cannot be produced. Eq.~(\ref{equation:A_et_B}b) shows
that $\det(\negr B)=0$ when for one leg $i$, $(\negr b_i - \negr
a_i) \perp (\negr c_i - \negr b_i)$, where $\negr a_i$ is the
position vector of $A_i$. Thus, the serial singularities form
three planes orthogonal to the $x$, $y$ and $z$ axis,
respectively.

It will be shown in Section~\ref{Section4} that the optimization
of the Orthoglide puts the serial and parallel singularities far
away from the Cartesian workspace. Also, even if the direct and
inverse kinematics may theoretically have several solutions, only
one solution exists in the Cartesian workspace \cite
{Wenger:2000}.
\section{Optimization of the Design Parameters}

The aim of this section is to define the geometric parameters of
the Orthoglide as a function of the size of a prescribed cubic
Cartesian workspace with bounded transmission factors. We first
show that the orthogonal arrangement of the linear joints is
imposed by the condition on the isotropy and manipulability: we
want the Orthoglide to have an isotropic configuration with
velocity and force transmission factors equal to one. Then, we
impose that the transmission factors remain under prescribed
bounds throughout the prescribed Cartesian workspace and we deduce
the link dimensions and the joint limits. Limiting the force and
velocity transmission factors makes it possible to guarantee a
minimal kinematic stiffness throughout the Cartesian workspace.
The structural stiffness ({\it i.e.} including the stiffness of
all rods) is guaranteed by the over-constrained design and
preliminary rods stiffness analyses \cite{Wenger:1999a}. A more
detailed study of the Othoglide structural stiffness is currently
conducted at IRCCyN with finite element analyses.

\subsection{Condition Number and Isotropic Configuration}
The Jacobian matrix is said to be isotropic when its condition
number attains its minimum value of one \cite{Wenger:2000}. The
condition number of the Jacobian matrix is an interesting
performance index which characterises the distortion of a unit
ball under the transformation represented by the Jacobian matrix.
The Jacobian matrix of a manipulator is used to relate (i) the
joint rates and the Cartesian velocities, and (ii) the static load
on the output link and the joint torques or forces. Thus, the
condition number of the Jacobian matrix can be used to measure the
uniformity of the distribution of the tool velocities and forces
in the Cartesian workspace.
\subsection{Isotropic Configuration of the Orthoglide}
For parallel manipulators, it is more convenient to study the
conditioning of the Jacobian matrix that is related to the inverse
transformation, $\negr J^{-1}$. When \negr B is not singular,
$\negr J^{-1}$ is defined by:
 \bed
   \dot{\gbf \rho} = \negr J^{-1} \dot{\negr p}
   {\rm ~~with~~}
   \negr J^{-1} = \negr B^{-1} \negr A
 \eed
 Thus:
 \beqa
   \negr J^{-1} =
   \left[\begin{array}{c}
            (1/\eta_1) (\negr c_1 - \negr b_1)^T \\
            (1/\eta_2) (\negr c_2 - \negr b_2)^T \\
            (1/\eta_3) (\negr c_3 - \negr b_3)^T
         \end{array}
   \right]
   \label{equation:J}
 \eeqa
with $\eta_i= (\negr c_i - \negr b_i)^T \negr n_i $ for $i =
1,2,3$.

The matrix $\negr J^{-1}$ is isotropic when $\negr J^{-1}\negr
J^{-T}=\sigma^2 \negr 1_{3 \times 3}$, where $\negr 1_{3 \times
3}$ is the $3 \times 3$ identity matrix. Thus, we must have,
 \bseq
   \be
     \frac{1}{\eta_1} ||\negr c_1 - \negr b_1|| =
     \frac{1}{\eta_2} ||\negr c_2 - \negr b_2|| =
     \frac{1}{\eta_3} ||\negr c_3 - \negr b_3||
   \ee
   \be
     (\negr c_1 - \negr b_1)^T
     (\negr c_2 - \negr b_2) = 0
   \ee
   \be
     (\negr c_2 - \negr b_2)^T
     (\negr c_3 - \negr b_3) = 0
   \ee
   \be
     (\negr c_3 - \negr b_3)^T
     (\negr c_1 - \negr b_1) = 0
   \ee
   \label{equation:isotropie}
 \eseq
Equation~(\ref{equation:isotropie}a) states that the angle between
the axis of the linear joint and the link $B_iC_i$ must be the
same for each leg $i$. Equations~(\ref{equation:isotropie}b--d)
mean that the links $B_iC_i$ must be orthogonal to each other.
Figure \ref{figure:Orthoglide_Workspace} shows the isotropic
configuration of the Orthoglide. Note that the orthogonal
arrangement of the linear joints is not a consequence of the
isotropy condition, but it stems from the condition on the
transmission factors at the isotropic configuration, as shown in
the next section.
\subsection{Transmission factors}
For serial 3-axis machine tools, a motion of an actuated joint
yields the same motion of the tool (the transmission factors are
equal to one). For parallel machines, these motions are generally
not equivalent. When the mechanism is close to a parallel
singularity, a small joint rate can generate a large velocity of
the tool. This means that the positioning accuracy of the tool is
lower in some directions for some configurations close to parallel
singularities because the encoder resolution is amplified. In
addition, a velocity amplification in one direction is equivalent
to a loss of stiffness in this direction.

The manipulability ellipsoids of the Jacobian matrix of robotic
manipulators was defined two decades ago \cite{Salisbury:1982}.
This concept has then been applied as a performance index to
parallel manipulators \cite{Kim:1997}. Note that, although the
concept of manipulability is close to the concept of condition
number, they do not provide the same information. The condition
number quantifies the proximity to an isotropic configuration,
{\it i.e.} where the manipulability ellipsoid is a sphere, or, in
other words, where the transmission factors are the same in all
the directions, but it does not inform about the value of the
transmission factor.

The manipulability ellipsoid of $\negr J^{-1}$ is used here for
(i) defining the orientation of the linear joints and (ii)
defining the joint limits of the Orthoglide such that the
transmission factors are bounded in the prescribed Cartesian
workspace.

We want the transmission factors to be equal to one at the
isotropic configuration like for a serial machine tool. This
condition implies that the three terms of
Eq.~(\ref{equation:isotropie}a) must be equal to one:
   \beqa
     \frac{1}{\eta_1} ||\negr c_1 - \negr b_1|| =
     \frac{1}{\eta_2} ||\negr c_2 - \negr b_2|| =
     \frac{1}{\eta_3} ||\negr c_3 - \negr b_3|| = 1
     \label{eqution:isotropie-1}
   \eeqa
which implies that $(\negr b_i - \negr a_i)$ and  $(\negr c_i -
\negr b_i)$ must be collinear for each $i$.

Since, at this isotropic configuration, links $B_iC_i$ are
orthogonal, Eq.~(\ref{eqution:isotropie-1}) implies that the links
$A_iB_i$ are orthogonal, {\it i.e.} the linear joints are
orthogonal. For joint rates belonging to a unit ball, namely,
$||\dot{\gbf \rho}|| \leq 1$, the Cartesian velocities belong to
an ellipsoid such that:
 \bed
   \dot{\gbf p}^T (\negr J \negr J^T) \dot{\gbf p} \leq 1
 \eed
The eigenvectors of matrix $(\negr J \negr J^T)^{-1}$ define the
direction of its principal axes of this ellipsoid and the square
roots $\xi_1$, $\xi_2$ and $\xi_3$ of the eigenvalues of $(\negr J
\negr J^T)^{-1}$ are the lengths of the aforementioned principal
axes. The velocity transmission factors in the directions of the
principal axes are defined by $\psi_1 = 1 / \xi_1$, $\psi_2 = 1 /
\xi_2$ and $\psi_3 = 1 / \xi_3$. To limit the variations of this
factor, we impose
 \be
   \psi_{min} \leq \psi_i \leq \psi_{max} 
   \label{e:velocity_limits}
 \ee
throughout the Cartesian workspace. This condition determines the
link lengths and the linear joint limits. To simplify the problem,
we set $\psi_{min}=1/ \psi_{max}$.
\subsection{Design of the Orthoglide for a Prescribed Cartesian Workspace}
\label{Section4}
\begin{figure}[!ht]
    \begin{center}
           \includegraphics[width=80mm,height=80mm]{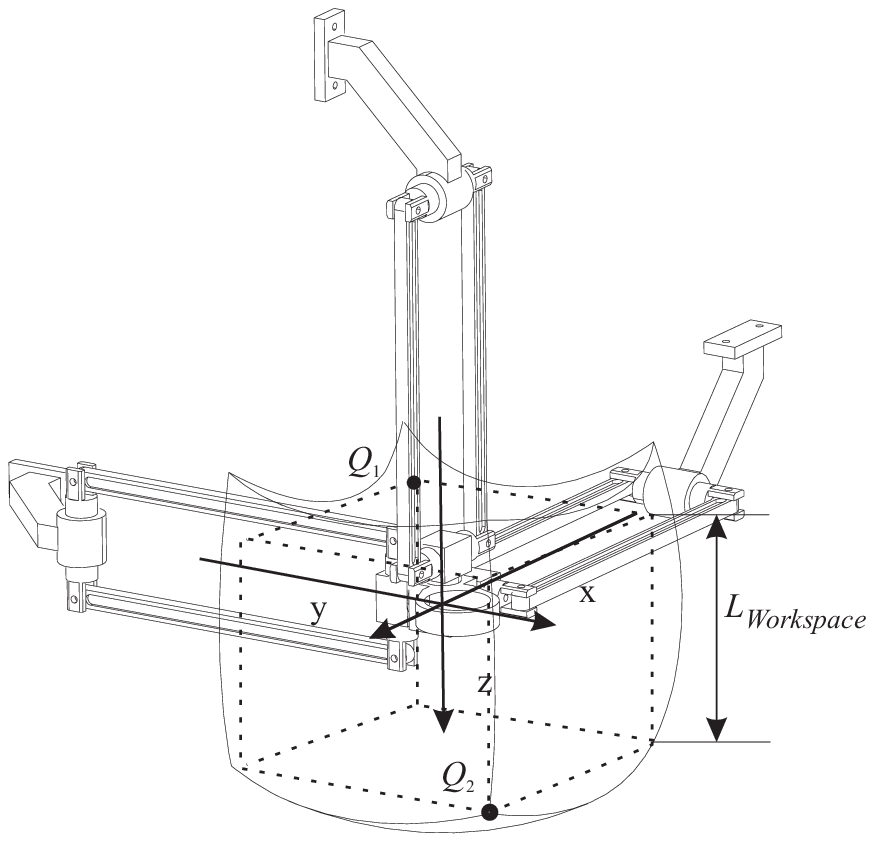}
           \caption{Isotropic configuration and Cartesian workspace of the Orthoglide mechanism and points $Q_1$ and $Q_2$}
           \protect\label{figure:Orthoglide_Workspace}
    \end{center}
 \end{figure}

For usual machine tools, the Cartesian workspace is generally
given as a function of the size of a right-angled parallelepiped.
Due to the symmetrical architecture of the Orthoglide, the
Cartesian workspace has a fairly regular shape. In fact, the
workspace is defined by the intersection of three orthogonal
cylinders topped with spheres. As shown in
Fig.~\ref{figure:Orthoglide_Workspace}, it is easy to include a
cube whose sides are parallel to the planes $xy$, $yz$ and $xz$
respectively.
The aim of this section is to define the position of the base
point $A_i$, the link lengths $L$ and the linear actuator range
$\Delta \rho$ with respect to the limits on the transmission
factors defined in Eq.~(\ref{e:velocity_limits}) and as a function
of the size of the prescribed Cartesian workspace $L_{Workspace}$.

The proposed optimization scheme is divided into three steps.
\begin{enumerate}
 \item
First, two points $Q_1$ and $Q_2$ are determined in the prescribed
cubic Cartesian workspace (Fig.~\ref{figure:Orthoglide_Workspace})
such that if the transmission factor bounds are satisfied at these
points, they are satisfied in all the prescribed Cartesian
workspace.
 \item
The points $Q_1$ and $Q_2$ are used to define the leg length $L$
as function of the size of the prescribed cubic Cartesian
workspace.
 \item
Finally, the positions of the base points $A_i$ and the linear
actuator range $\Delta \rho $ are calculated such that the
prescribed cubic Cartesian workspace is fully included in the
Cartesian workspace of the Orthoglide.
\end{enumerate}

\begin{figure}[!ht]
    \begin{center}
           \includegraphics[width=65mm,height=62mm]{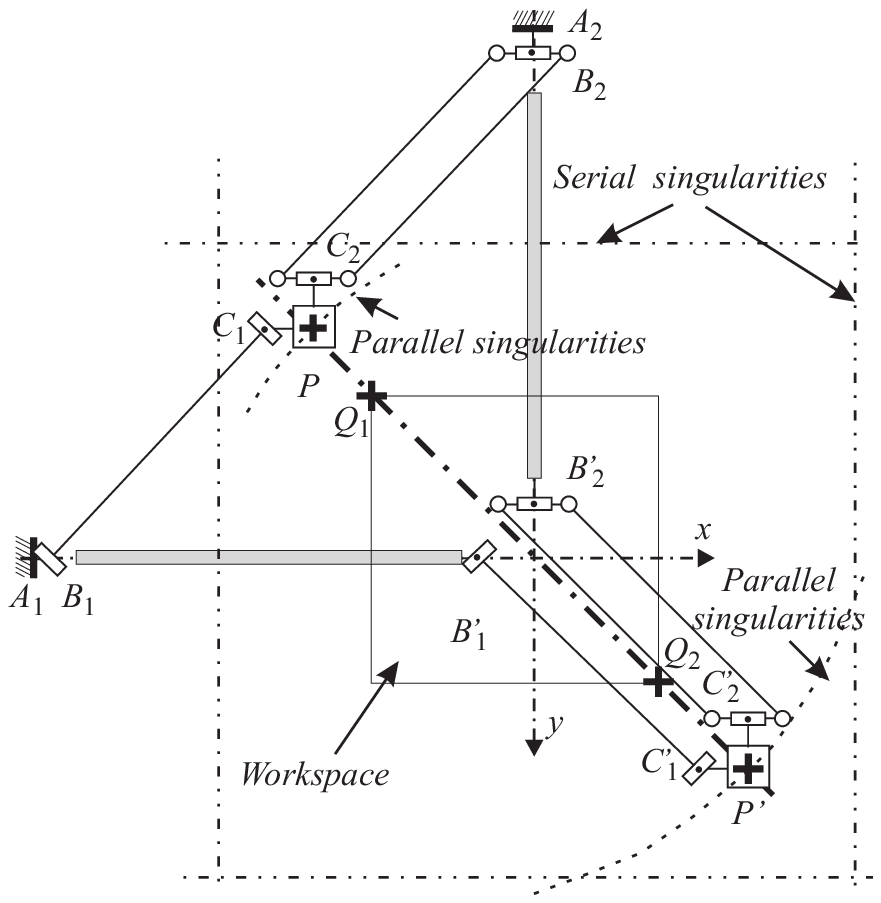}
           \caption{Points $Q_1$ and $Q_2$ and the singular configurations (top view)}
           \protect\label{figure:Workspace_Singularities2}
    \end{center}
\end{figure}

{\bf Step~1:} The transmission factors are equal to one at the
isotropic configuration. These factors increase or decrease when
the tool center point moves away from the isotropic configuration
and they tend towards zero or infinity in the vicinity of the
singularity surfaces. It turns out that the points $Q_1$ and $Q_2$
defined at the intersection of the Cartesian workspace boundary
with the axis $x = y = z$ (in a reference frame (O, $x$, $y$, $z$)
centered at the intersection of the three linear joint axes,
Fig.~\ref{figure:Orthoglide_Workspace}) are the closest ones to
the singularity surfaces, as illustrated in
Fig.~\ref{figure:Workspace_Singularities2} which shows on the same
top view the Orthoglide in the two parallel singular
configurations of figures~\ref{figure:Coplanarsing} and
~\ref{figure:Parallelsing}. Thus, we may postulate the intuitive
result that if the prescribed bounds on the transmission factors
are satisfied at $Q_1$ and $Q_2$, then these bounds are satisfied
throughout the prescribed cubic Cartesian workspace. In fact, this
result can be proved using interval analysis \cite{Chablat:2002}.

\begin{figure}[!ht]
    \begin{center}
           \includegraphics[width=65mm,height=62mm]{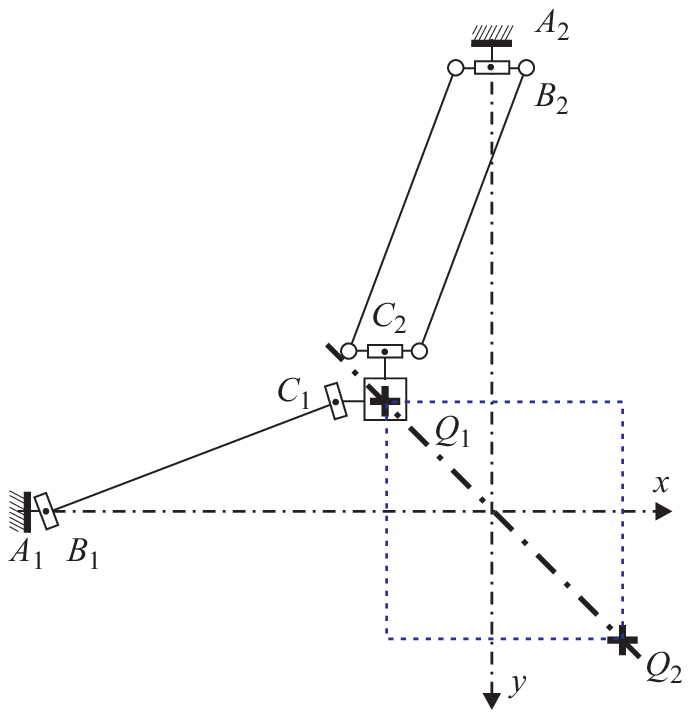}
           \caption{$Q_1$ configuration}
           \protect\label{figure:Workspace1}
    \end{center}
 \end{figure}

\begin{figure}[!ht]
    \begin{center}
           \includegraphics[width=65mm,height=62mm]{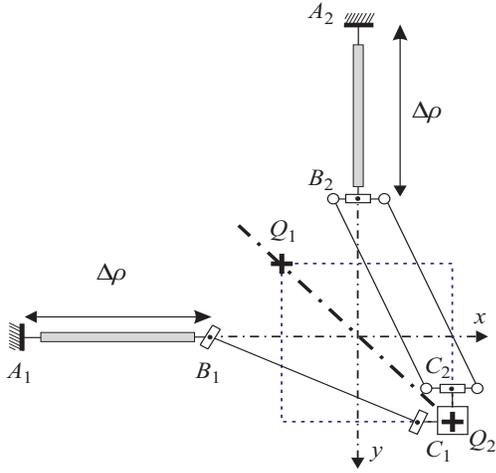}
           \caption{$Q_2$ configuration}
           \protect\label{figure:Workspace2}
    \end{center}
 \end{figure}

{\bf Step~2:} At the isotropic configuration, the angles
$\theta_i$ and $\beta_i$ are equal to zero by definition. When the
tool center point $P$ is at $Q_1$,
$\rho_1=\rho_2=\rho_3=\rho_{min}$ (Fig.~\ref{figure:Workspace1}).
When $P$ is at $Q_2$, $\rho_1=\rho_2=\rho_3=\rho_{max}$
(Fig.~\ref{figure:Workspace2}).

We pose $\rho_{min}=0$ for more simplicity.

The position of $P$ along the $z$ axis can be written equivalently
as $z = - \sin(\beta_1) L$ and $z = \sin(\theta_2) \cos(\beta_2)
L$ by traversing the two chains $A_1B_1C_2P$ and $A_2B_2C_2P$,
respectively. On the axis $(Q_1Q_2)$, $\beta_1 = \beta_2 =
\beta_3$ and $\theta_1= \theta_2= \theta_3$. We note,
 \beqa
   \beta_1 = \beta_2 = \beta_3 = \beta \quad
   {\rm and} \quad
   \theta_1= \theta_2= \theta_3= \theta
   \label{e:beta_123}
 \eeqa

Thus, the angle $\beta$ can be written as a function of $\theta$,
 \beqa
   \beta= -\arctan(\sin(\theta))
   \label{e:beta_theta}
 \eeqa
Finally, by substituting Eq.~(\ref{e:beta_theta}) into
Eq.~(\ref{equation:J}), the inverse Jacobian matrix $\negr J^{-1}$
can be simplified as follows
 \bed
 \negr  J^{-1}=
   \left[\begin {array}{ccc}
       1             &
       -\tan(\theta) &
       -\tan(\theta)\\
       -\tan(\theta) &
       1             &
       -\tan(\theta) \\
       -\tan(\theta) &
       -\tan(\theta) &
       1
       \end{array}
   \right]
 \eed
Thus, the square roots of the eigenvalues of $(\negr J \negr
J^T)^{-1}$ are,
 \bed
   \xi_1= |2 \tan(\theta)-1| \quad {\rm and} \quad
   \xi_2= \xi_3= |\tan(\theta)+1|
 \eed
And the three velocity transmission factors are,
 \be
   \psi_1= \frac{1}{|2 \tan(\theta)-1|}  \quad {\rm and} \quad
   \psi_2= \psi_3= \frac{1}{|\tan(\theta)+1|}
   \label{e:frontieres}
 \ee

The joint limits on $\theta$ are located on both sides of the
isotropic configuration. To calculate the joint limits, we solve
the following inequations,
 \bseq
 \beqa
    \frac{1}{\psi_{max}} \leq \frac{1}{|2 \tan(\theta)-1|} \leq \psi_{max} \\
    \frac{1}{\psi_{max}} \leq \frac{1}{|\tan(\theta)+1|} \leq \psi_{max}
 \eeqa
 \label{e:constraints}
 \eseq
where the value of $\psi_{max}$ depends on the performance
requirements. Two sets of joint limits
($[\theta_{Q_1}~\beta_{Q_1}]$ and $[\theta_{Q_2}~\beta_{Q_2}]$)
are found in symbolic form. The detail of this calculation is
given in the Appendix.

The position vectors $\negr q_1$ and $\negr q_2$ of the points
$Q_1$ and $Q_2$, respectively, can be easily defined as a function
of $L$ (Figs.~\ref{figure:Workspace1} and
\ref{figure:Workspace2}),
 \bseq
 \beq
    \negr q_1= [q_1~q_1~q_1]^T
    \quad {\rm and} \quad
    \negr q_2= [q_2~q_2~q_2]^T
   \label{e:q1-q2}
 \eeq
with
 \beq
    q_1= - \sin(\beta_{Q_1}) L
    \quad {\rm and} \quad
    q_2= - \sin(\beta_{Q_2}) L
 \eeq
 \eseq

\begin{figure}[!ht]
    \begin{center}
           \includegraphics[width=60mm,height=57mm]{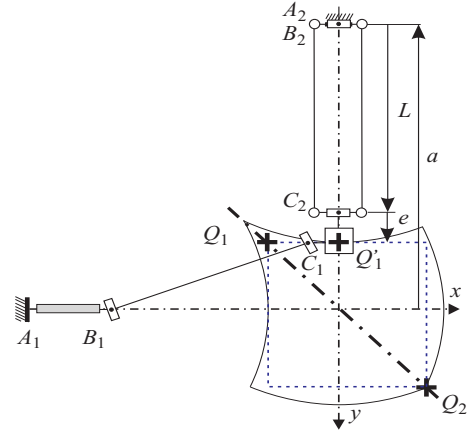}
           \caption{The point $Q_1'$ used for the determination of $a$}
           \protect\label{figure:Workspace3}
    \end{center}
\end{figure}

The size of the Cartesian workspace is,
 \bed
   L_{Workspace} = |q_2 - q_1|
 \eed
Thus, $L$ can be defined as a function of $L_{Workspace}$.
  \bed
    L= \frac{L_{Workspace}}{|\sin(\beta_{Q_2}) - \sin(\beta_{Q_1}) |}
  \eed
{\bf Step~3:} We want to determine the positions of the base
points, namely, $a=OA_1=OA_2=OA_3$. When the tool center point P
is at $Q_1'$ defined as the projection onto the $y$ axis of $Q_1$,
$\rho_2=0$ and, (Fig.~\ref{figure:Workspace3}) \beqa
    OA_2=OQ_1'+Q_1'C_2+C_2A_2 \nonumber
\eeqa
 Since $\rho_2=0$, $C_2A_2=C_2B_2=L$. With $OA_2=a$, $Q_1'C_2=PC_2=-e$
 and $OQ_1'=q_1$, we get,
 \bed
   a= q_1 -  e - L
 \eed

Since $q_1$ is known from Eqs.~(\ref{e:q1-q2}) and
~(\ref{e:joint-limit1}), $a$ can be calculated as function of $e$,
$L$ and $\psi_{max}$.

Now, we have to calculate the linear joint range $\Delta
\rho=\rho_{max}$ (we have posed $\rho_{min}$=0).

When the tool center point $P$ is at $Q_2$, $\rho=\rho_{max}$.
Projecting $A_2P=A_2B_2+B_2C_2+C_2P$ on the $y$ axis yields,
 \bed
    \rho_{max} = q_2 - a - \cos(\theta_{Q_2}) \cos(\beta_{Q_2}) L - e
 \eed
\subsection{Prototype}
\begin{figure}[!ht]
  \begin{center}
    \includegraphics[width=41mm,height=50mm]{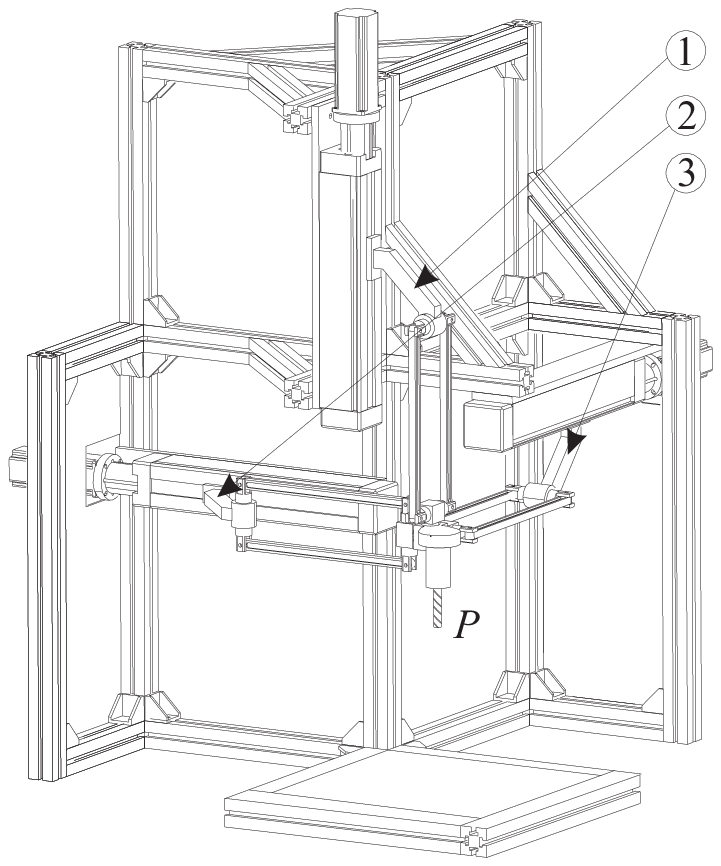}
    \includegraphics[width=41mm,height=50mm]{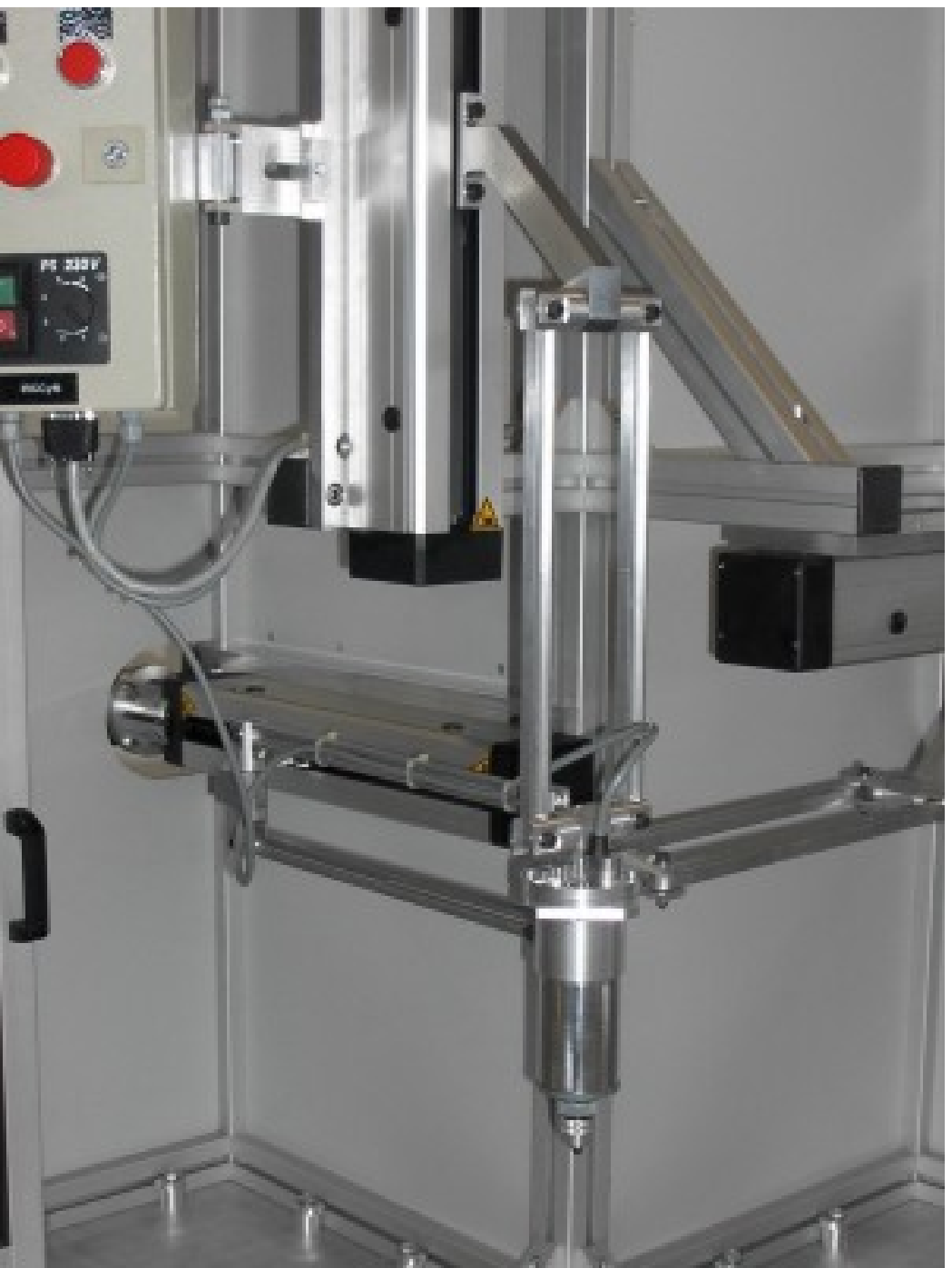}
    \caption{Catia model of the Orthoglide (left) and prototype (right)}
    \protect\label{figure:Prototype}
  \end{center}
\end{figure}

\begin{figure}[!ht]
    \begin{center}
        \includegraphics[width=87mm,height=30mm]{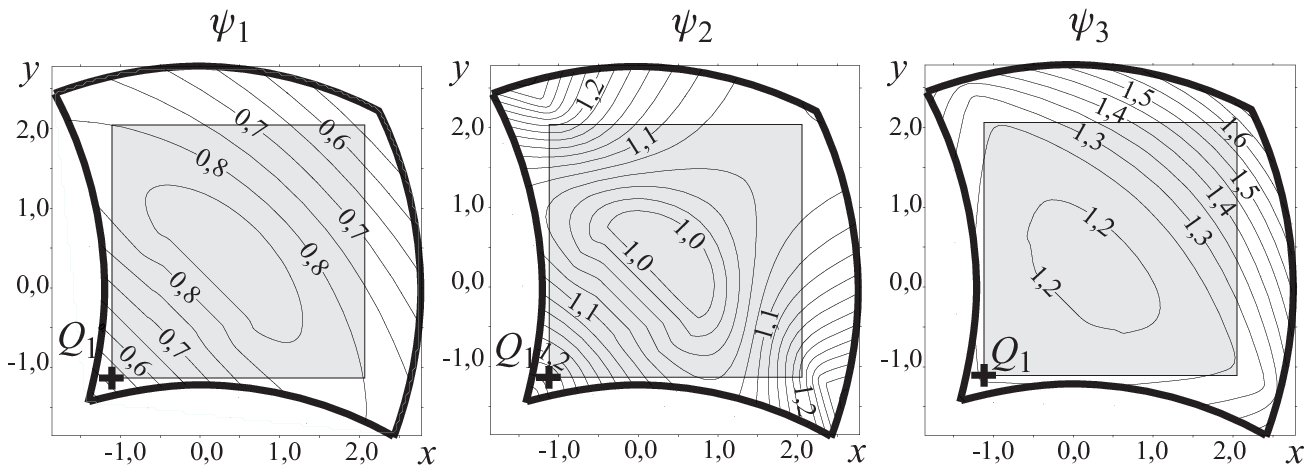}
           \caption{The three velocity transmission factors in a
           $z$-cross section of the Cartesian workspace passing through $Q_1$}
           \protect\label{figure:Isovalues_1}
    \end{center}
\end{figure}

Using the aforementioned two kinetostatic criteria, a small-scale
prototype has been constructed in our laboratory (
Figure~\ref{figure:Prototype} ).
The three parts (1), (2) and (3) have been designed to prevent
each parallelogram from colliding with the corresponding linear
motion guide. Also, the shifted position of the tool center point
$P$ limits the collisions between the parallelograms and the
workpiece. The actuated joints used for this prototype are rotary
motors with ball screws. The prescribed performances of the
Orthoglide prototype are a Cartesian velocity of $1.2 m/s$ and an
acceleration of $14m/s^2$ at the isotropic point. The desired
payload is $4 kg$. The size of its prescribed cubic Cartesian
workspace is $200 \times 200 \times 200~mm$. We limit the
variations of the velocity transmission factors as,
 \be
   1/2 \leq \psi_i \leq 2
   \label{e:velocity_limits_prototype}
 \ee
The resulting length of the three parallelograms is $L=310~mm$ and
the resulting range of the linear joints is $\Delta~\rho=~
257~mm$. Thus, the ratio of the range of the actuated joints to
the size of the prescribed Cartesian workspace is
$r=200/257=0.78$. This ratio is high compared to other PKM. The
three velocity transmission factors are depicted in
Fig.~\ref{figure:Isovalues_1}. These factors are given in a
$z$-cross section of the Cartesian workspace passing through
$Q_1$.

\section{Conclusions}
The Orthoglide is a new Delta-type PKM dedicated to 3-axis rapid
machining applications that was designed to meet the advantages of
both serial 3-axis machines (regular workspace and homogeneous
performances) and parallel kinematic architectures (good dynamic
performances). A systematic procedure has been provided to define
the geometric parameters of the Orthoglide as functions of the
size of a prescribed cubic Cartesian workspace and bounded
velocity and force transmission factors.

The Orthoglide has been designed under isotropic conditions and
limited transmission factors. Low inertia and intrinsic stiffness
have been set as additional design requirements. Thus, three
articulated parallelograms have been used, rather than legs
subject to bending as in the fully isotropic mechanisms proposed
in \cite{Carricato:2002,Kong:2002,Kim:2002}. At the isotropic
configuration, a displacement of a linear joint yields the same
displacement of the tool in the corresponding Cartesian direction
like in a serial machine. The Cartesian workspace is simple,
regular and free of singularities and self-collisions. It is
fairly regular and the performances are homogeneous throughout the
Cartesian workspace. Thus, the entire Cartesian workspace is
really available for tool paths. These features make the
Orthoglide a novel design as compared to the existing Delta-type
PKM structures. A small-scale prototype Orthoglide has been built
at IRCCyN to demonstrate the feasibility of the design. Dynamic
model based control laws will be implemented \cite{Guegan:2002}
and first machining experiments with plastic parts will be
conducted.
\section*{Acknowledgments}
This works is supported by R\'egion Pays-de-Loire, Agence
Nationale pour la Valorisation de la Recherche, \'Ecole des Mines
de Nantes and C.N.R.S.

S. Bellavoir, G. Branchu, P. Lemoine and P. Molina are gratefully
acknowledged for their technical help.
\bibliographystyle{unsrt}

\section{Appendix}
To calculate the joint limits on $\theta$ and $\beta$, we solve
the followings inequations, from the Eqs.~\ref{e:constraints},
  \be
    |2\tan(\theta)-1| \leq \psi_{max} \quad
    \frac{1}{|2\tan(\theta)-1|} \leq \psi_{max}
  \ee
Thus, we note,
 \be
   f_1= |2\tan(\theta)-1| \quad
   f_2= 1/|2\tan(\theta)-1|
 \ee
\begin{figure}[!ht]
    \begin{center}
           \includegraphics[width=71mm,height=43mm]{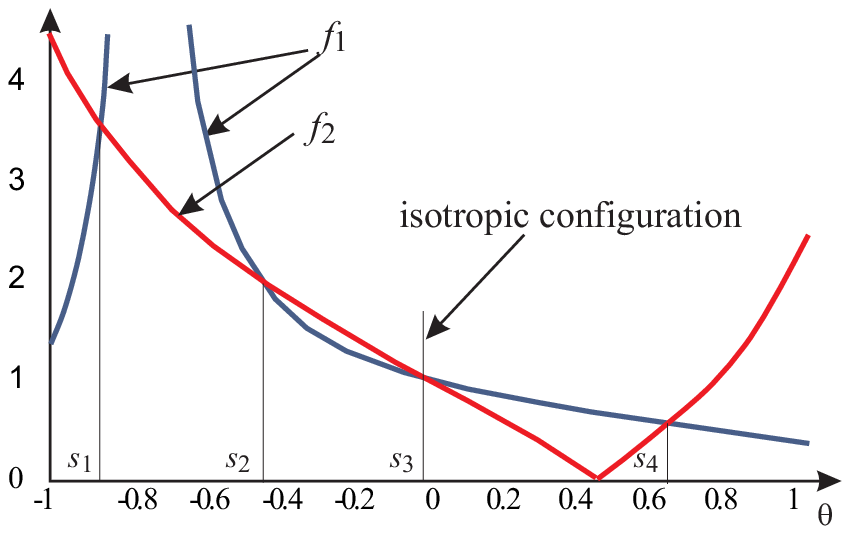}
           \caption{$f_1$ and $f_2$ as function of $\theta$ along $(Q_1Q_2)$}
           \protect\label{figure:variations_theta}
    \end{center}
\end{figure}
Fig.~(\ref{figure:variations_theta}) shows $f_1$ and $f_2$ as
function of $\theta$ along $(Q_1Q_2)$. The four roots of $f_1=f_2$
in $[-\pi~\pi]$ are,
 \bseq
 \beqa
   s_1&=& -\arctan\left((1+\sqrt{17})/4\right) \\
   s_2&=& -\arctan\left(1/2\right) \\
   s_3&=& 0 \\
   s_4&=& \arctan\left((-1+\sqrt{17})/4\right)
 \eeqa
with
 \beqa
   f_1(s_1)= (-3+\sqrt{17})/4 \quad
   f_1(s_2)= 2\\
   f_1(s_3)= 1 \quad
   f_1(s_4)= (3+\sqrt{17})/4
 \eeqa
 \eseq
The isotropic configuration is located at the configuration where
$\theta=\beta=0$. The limits on $\theta$ and $\beta$ are in the
vicinity of this configuration. Along the axis $(Q_1Q_2)$, the
angle $\theta$ is lower than $0$ when it is close to $Q_2$, and
greater than $0$ when it is close to $Q_1$.

To find $\theta_{Q_1}$, we study the functions $f_1$ and $f_2$
which are both decreasing on $[0~\arctan(1/2)]$. Thus, we have,
 \bseq
  \beqa
     \theta_{Q_1} &=&  \arctan\left({\frac {{\psi_{max}}-1}{{2\psi_{max}}}}\right)
     \\
     \beta_{Q_1}  &=& -\arctan\left({\frac {\psi_{max}-1}{\sqrt {5\psi_{max}^{2}-2\psi_{max}+1}}}\right)
     \label{e:joint-limit1}
  \eeqa
 \eseq
In the same way, to find $\theta_{Q_2}$, we study the functions
$f_1$ and $f_2$ on $[s_1~0]$. The three roots $s_1$, $s_2$ and
$s_3$ define two intervals. If $\psi_{max} \in
[f_1(s_1)~f_1(s_2)]$, we have,
 \bseq
  \beqa
     \theta_{Q_2} &=& -\arctan\left({\frac {{\psi_{max}}-1}{{\psi_{max}}}}\right)
     \label{e:joint-limit2} \\
     \beta_{Q_2}  &=&  \arctan\left({\frac {\psi_{max}-1}{\sqrt {2\psi_{max}^{2}-2\psi_{max}+1}}}\right)
  \eeqa
otherwise, if $\psi_{max} \in [f_1(s_2)~f_1(s_3)]$,
  \beqa
     \theta_{Q_2} &=& -\arctan\left({\frac {\psi_{max}-1}{{2}}}\right)
     \label{e:joint-limit3}  \\
     \beta_{Q_2}  &=&  \arctan\left({\frac {\psi_{max}-1}{\sqrt {{\psi_{max}}^{2}-2\psi_{max}+5}}}\right)
  \eeqa
 \eseq
\end{document}